\pdfoutput=1

\documentclass[11pt]{article}
\usepackage[square, comma, sort&compress, numbers]{natbib}
\usepackage{ACL2023}
\usepackage{graphicx}
\usepackage{times}
\usepackage{latexsym}
\usepackage{algorithm}
\usepackage{algorithmic}
\usepackage{amsmath,amsfonts}
\usepackage{multirow}
\usepackage{booktabs}
\usepackage[T1]{fontenc}

\usepackage{microtype}

\usepackage[utf8]{inputenc}

\usepackage{microtype}

\usepackage{inconsolata}
\title{DialogueLLM: Context and Emotion Knowledge-Tuned Large Language Models for Emotion Recognition in Conversations}

\author{Yazhou Zhang$^{1,2}$, \ Mengyao Wang$^{2}$, Youxi Wu$^3$\Thanks{Corresponding authors}, \ Prayag Tiwari$^4$, \ Qiuchi Li$^5$, \ Benyou Wang$^6$, \ Jing Qin$^{1*}$\\
    $^1$The Hong Kong Polytechnic University \\
    $^2$Zhengzhou University of Light Industry\\ 
    $^3$Hebei University of Technology\\
    $^4$Halmstad University\\ 
    $^5$Copenhagen University \\
    $^6$The Chinese University of Hong Kong, Shenzhen \\
	}     

\begin{document}
\maketitle
\begin{abstract}
Large language models (LLMs) and their variants have shown extraordinary efficacy across numerous downstream natural language processing (NLP) tasks, which has presented a new vision for the development of NLP. Despite their remarkable performance in natural language generating, LLMs lack a distinct focus on the emotion understanding domain. As a result, using LLMs for emotion recognition may lead to suboptimal and inadequate precision. Another limitation of the current emotion LLMs is that they are typical trained without leveraging multi-modal information. To overcome these limitations, we propose DialogueLLM, a context and emotion knowledge tuned LLM that is obtained by fine-tuning large language models with benchmarking multi-modal (i.e., texts and videos) emotional dialogues. The visual information is considered as the supplementary knowledge to construct high-quality instructions. We offer a comprehensive evaluation of our proposed model on three benchmarking emotion recognition in conversations (ERC) datasets and compare the results against the state-of-the-art baselines and other state-of-the-art LLMs. Additionally, DialogueLLM-7B can be easily trained using LoRA on a 40GB A100 GPU in 5 hours, facilitating reproducibility for other researchers. 
\end{abstract}

\section{Introduction}

Scaling up language models has been proved to be an effective way to improve the performance and sample efficiency in downstream NLP tasks. The rise of instruction-following LLMs has garnered considerable attention from academy and industry, due to their outstanding performance in  human instruction understanding and responsing. Language modeling has evolved from small language models (SLMs), e.g., GPT~\cite{radford2018improving}, BERT~\cite{devlin-etal-2019-bert}, RoBERTa~\cite{liu2019roberta}, etc., to LLMs, e.g., ChatGPT\footnote{https://chat.openai.com/} GPT-4~\cite{openai2023gpt4}, Claude\footnote{https://www.anthropic.com/product}, etc. 
 \begin{figure}[t]
    \centering
    \includegraphics[height=3.13in, width=3.1in]{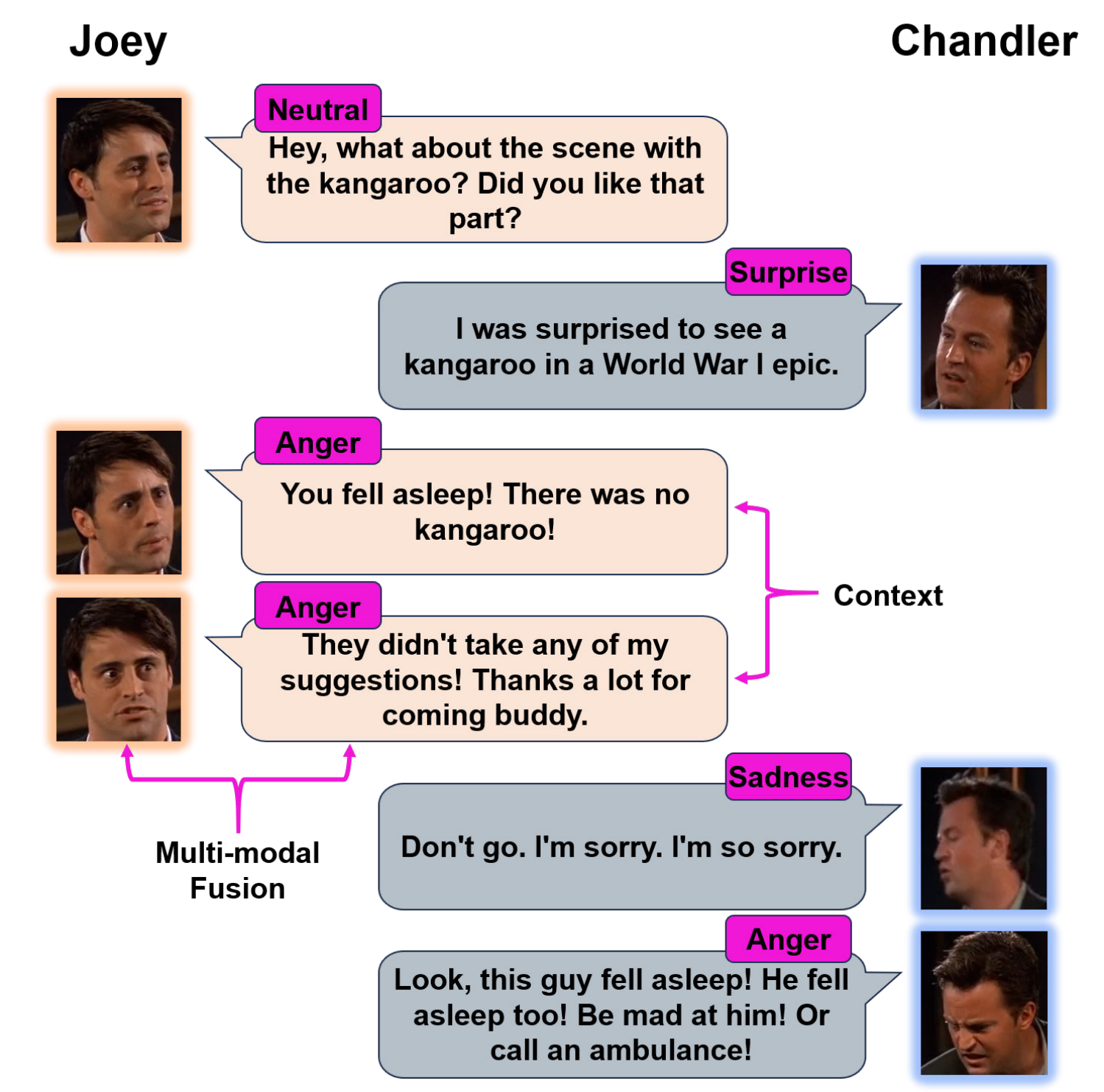}
  \caption{Sample utterances in a multi-modal conversation from the MELD dataset.}
   \label{fig:fig1}
\end{figure}

Compared with SLMs, LLMs are characterized by their enormous parameter size, typically reaching tens of billions or even more. They often have stronger generalization across various downstream tasks and unique emergent ability to tackle complex tasks. Despite that LLMs possess numerous commendable qualities, they also present a couple of limitations that deserve careful consideration and in-depth exploration: (1) the non-open source status may restrict the development of LLMs community and (2) they are not specifically designed for emotion understanding task. Their broad domain knowledge frequently proves insufficient when tackling such specialized domains. For example, Zhang et al.~\cite{zhang2023sentiment} showed LLMs' unsatisfactory performance in many emotion recognition tasks without fine-tunning on emotional knowledge. Let et al.~\cite{DBLP:InstructERC} presented a retreival based framework to improve the adaptability of LLMs to emotion recognition.
Hence, the potential of LLMs in understanding emotional communication needs to be explored further.

Human communication is the process of exchanging information, thoughts, ideas, and feelings between individuals, which is naturally filled with the participant's subjective attitudes or emotions. Emotion recognition in conversations (ERC) aims to accurately detect the feelings and emotions expressed in the utterances. It has immense potential in dialogue understanding and intent analysis, and has been an active task in the recent literature~\cite{10131957,zhang2023multitask,zhang2023m3gat}. In general, there are two key factors that contribute the classification performance, i.e., multi-modal fusion and context dependency (also known as intra- and inter-speaker dependency)~\cite{ma2023moving}. Multi-modal fusion involves combining information from different sources or modalities, such as text, visual cues, to obtain a more comprehensive and accurate understanding of the emotional utterance. In view that emotions are influenced by the surrounding environment, the relationship between the participants, etc., context is a critical factor in accurately classifying emotions in conversations. The same utterance in different contexts might express different emotions. Fig.~\ref{fig:fig1} illustrates an example to introduce the presence of both challenges.

To overcome the above-mentioned limitations, it's crucial to develop emotion-tailored LLMs that can better understand human-human conversation and take a further step towards emotion intelligence. In this paper, we present DialogueLLM, an emotion and context knowledge enhanced language model, which is specifically designed for ERC based on the open-source base models, namely LLaMA 2~\cite{touvron2023llama2}. By collecting diverse instruction conversational data based on emotional knowledge from five open-source benchmarking datasets (i.e., MELD~\cite{poria2018meld}, IEMOCAP~\cite{busso2008iemocap}, EmoryNLP~\cite{zahiri2017emotion}.), we obtain 2411 multi-party dialogues, over 24,304 utterances. Meanwhile, the visual information (i.e., videos) will be forward into ERNIE Bot\footnote{https://yiyan.baidu.com/} to automatically generate the text descriptions, which will be considered as the supplementary knowledge to construct high-quality instructions. We adopt an end-to-end supervised instruction-finetuning approach on the open-source LLaMA 2-7B base models. Additionally, DialogueLLM-7B can be easily trained using LoRA on a 40GB A100 GPU in 5 hours, facilitating reproducibility for other researchers.

We offer a comprehensive evaluation of our proposed DialogueLLM model across three ERC tasks and compare the results against 15 state-of-the-art ERC baselines, including bc-LSTM~\cite{zhang2023learning}, MTL~\cite{Li2020MultiTaskLW}, ICON~\cite{hazarika2018icon}, DialogXL~\cite{shen2021dialogxl}, TODKAT~\cite{zhu-etal-2021-topic}, CoG-BART~\cite{li2022contrast}, DialogueGCN~\cite{ghosal2019dialoguegcn}, RGAT~\cite{ishiwatari2020relation}, DAG-ERC~\cite{shen2021directed}, DialogueRNN~\cite{majumder2019dialoguernn}, DialogueCRN~\cite{hu2021dialoguecrn}, CauAIN~\cite{zhao2022cauain}, COIN~\cite{zhang2021coin}, GraphCFC~\cite{DBLP:journals/corr/abs-2207-12261}, SACL-LSTM~\cite{hu2023supervised} and three SOTA LLMs, i.e., LLaMA, Alpaca\footnote{https://crfm.stanford.edu/2023/03/13/alpaca.html} and LLaMA 2.
The experimental results show the effectiveness of DialogueLLM with the margin of  5.36\%,  1.03\% and 1.5\% for three benchmarking ERC tasks. The study reveals that DialogueLLM significantly outperforms the SOTA baselines on ERC tasks requiring deeper understanding or conversational emotion information. A series of sub-experiments underscore how emotion and context knowledge enhanced LLMs deal with ERC tasks. The main innovations of the work are concluded as follows:

\begin{itemize}
\item To the best of our knowledge, DialogueLLM is the first open source emotional LLM that is specifically designed for ERC tasks.

\item The visual information is proposed to construct high-quality instructions.

\item We show a comprehensive dataset of over 24K utterances to serve as a knowledge corpus, supporting the training and testing of emotional LLMs with accurate and domain-specific knowledge.

\item Our model achieves state-of-the-art performance on ERC tasks. We show that an open-sourced model finetuned with emotional knowledge has the potential to achieve even higher accuracy than SOTA.
\end{itemize}

The rest of this paper is organized as follows. Section~\ref{sec:related} briefly outlines the related work. In Section~\ref{sec:model}, we describe the proposed DialogueLLM in detail. In Section~\ref{sec:experiments}, we
report the empirical experiments and analyze the results. Section~\ref{sec:conclusions} concludes the paper and points out future research directions.

\section{Related Work}\label{sec:related}

We depict two lines of research that form the basis of this work: large language models and emotion recognition in conversations models.

\subsection{Large Language Models}
In recent years, significant advancements in natural language processing (NLP) have been attributed to the emergence of large language models. These models have showcased remarkable capabilities such as in-context learning, few-shot prompting, instruction following, etc. These dynamic abilities have greatly contributed to boosting the effectiveness of language models, thus enabling AI algorithms to achieve unparalleled levels of effectiveness and productivity. Typically, models like the transformer architecture-based LLMs are first pre-trained using extensive datasets comprising diverse languages and domains~\cite{zhao2023survey}. 

OpenAI has achieved significant milestones with the creation of two groundbreaking models: ChatGPT and GPT-4. These models herald a new era in language processing. However, due to their proprietary nature, there has been a proliferation of LLM variants featuring tens or even hundreds of billions of parameters. Our aim is to categorize these LLMs into two distinct groups based on their specialization: general LLMs and specialized LLMs. General LLMs are designed for versatility across a wide spectrum of NLP tasks, including machine translation, language comprehension, and dialogue generation. Prominent examples of these models are GPT-4, Claude, ChatGPT, LLaMA, PanGu-$\Sigma$~\cite{ren2023pangu}, Bard~\footnote{https://bard.google.com/}, Falcon~\cite{penedo2023refinedweb}, etc. Such LLMs are not specifically optimized for any particular task. While they can perform well across a range of tasks, but their potentials in specific scenarios await further explore. 

In contrast, specialized LLMs also known as task-specific LLMs, are fine-tuned for specific tasks via task-specific architectures and knowledge, allowing them to achieve higher or comparable performance against general LLMs with fewer parameters. For example, Wang et al.~\cite{chen2023phoenix} released a large language model `Phoenix' to meet the needs of multiple languages. Liu and Low~\cite{liu2023goat} fine-tuned a Goat model based on LLaMA model to deal with arithmetic tasks. In view that LLMs have not yet performed optimally in medical domain tasks, a few Chinese and English medical knowledge enhanced LLMs have been proposed, such as HuaTuo~\cite{wang2023huatuo}, PMC-LLaMA~\cite{wu2023pmc}, Dr. LLaMA~\cite{guo2023dr}, ChatDoctor~\cite{li2023chatdoctor}. Different from the above-mentioned works, we aim to explore the potential of LLMs in emotion understanding domain and take a further step towards emotional intelligence.

 \begin{figure*}[t]
    \centering
  \includegraphics[height=2.8in, width=6.2in]{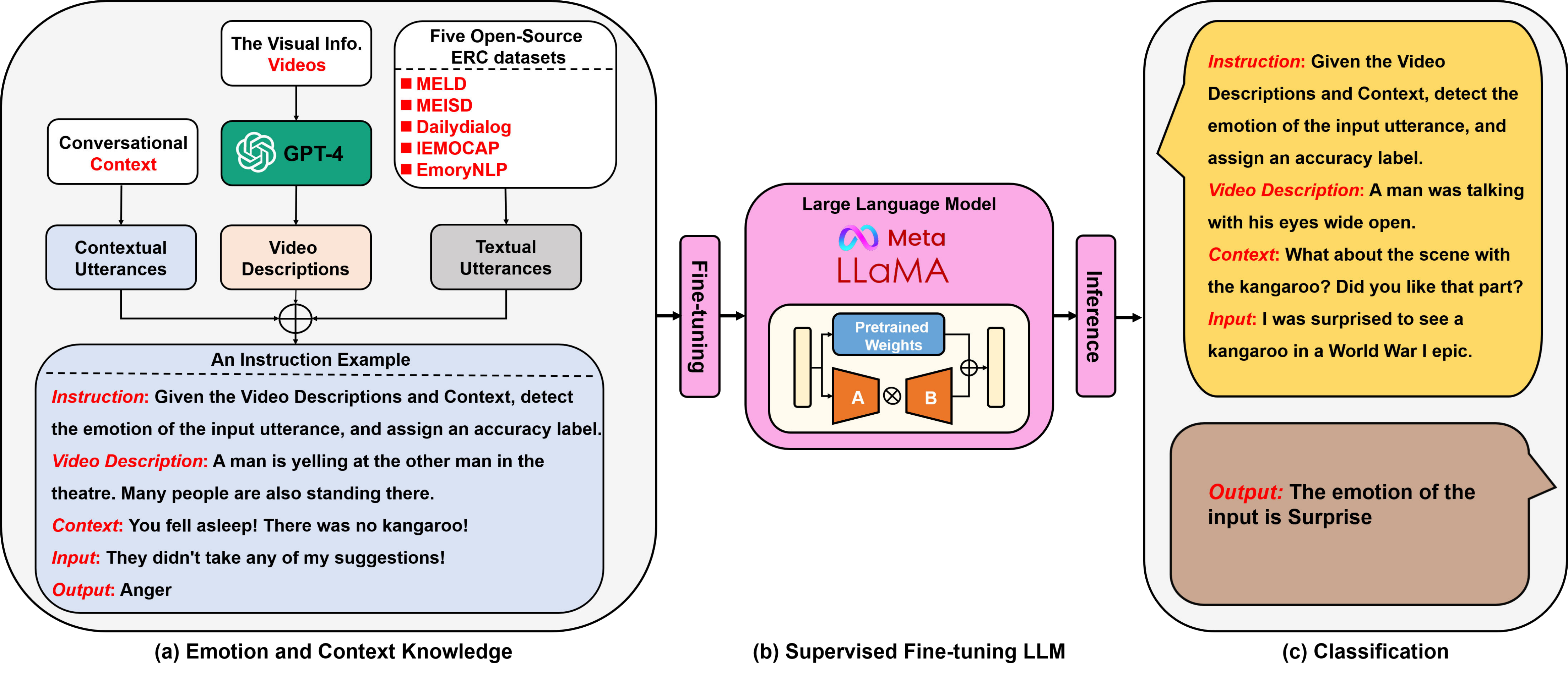}
  \caption{Overview of DialogueLLM fine-tuning and classification pipeline.}
   \label{fig:model}
\end{figure*}

\subsection{Emotion Recognition in Conversations}
Emotion recognition in conversation (ERC) has become a popular research topic. In this task, the conversational context dependency and multi-modal fusion have been considered through deep learning approaches. These efforts can be broadly categorized into methods based on sequences and those based on the Transformer architecture.

Sequence based approaches often use the sequential information in a dialogue to capture the contextual and emotional features. For example, Poria et al. \cite{poria2017context} introduced an LSTM-based model that effectively captured conversational context from surrounding videos, thereby enhancing the classification process. Building upon this idea, Hazarika et al. \cite{hazarika2018conversational} presented the conversational memory network (CMN), which harnessed contextual information from the conversation history to improve ERC. Another approach by Majumder et al. \cite{majumder2019dialoguernn} introduced the DialogueRNN model, which meticulously monitored the states of individual participants throughout the conversation, utilizing this information for ERC. In terms of multimodal advancements, Poria et al. \cite{poria2018meld} played a pivotal role by crafting the first-ever multimodal conversational dataset named the multimodal emotionlines dataset (MELD). This dataset was instrumental in propelling the field of conversational sentiment analysis. Further innovation came from Zhang et al. \cite{ijcai2019-Zhang}, who devised the quantum-inspired interactive network (QIN) model for conversational emotion recognition, showcasing its effectiveness. Moreover, their research extended to the realm of multi-task learning. Zhang et al. \cite{zhang2021cfn} devised a quantum-inspired multi-task learning framework catering to both sarcasm detection and emotion recognition in conversations. 

Transformer based approaches often adopt the ``fine-tuning'' paradigm. They build the models upon the foundation of Transformer based pre-trained language models. Then, such models are supervised-fine-tuned with labeled samples and are adapted to the specific task. For instance, Li et al.~\cite{li2022contrast} used a supervised contrastive term and a response generation task to enhance BART's ability for ERC. Zhang et al.~\cite{zhang2023m3gat} proposed a multi-modal multi-task network based on BERT and graph attention network (GAT) to detect sentiment and emotion. They also proposed a quantum inspired multi-task interactive Transformer to model sentiment and emotion~\cite{10131957}. Chudasama et al.~\cite{chudasama2022m2fnet} presented a multi-modal fusion network (M2FNet) to learn emotion-relevant multi-modal features by revising the Transformer encoder. Qiao et al.~\cite{qiao2023mutual} built a mutual-enhanced incongruity learning network upon RoBERTa and graph convolution networks to identify sarcasm. Pramanick~\cite{pramanick2022multimodal} combined self-attention with BERT to model intra-modal correspondence and optimal transport for cross-modal correspondence, aiming to discover sarcasm and humor. Lei et al.~\cite{DBLP:InstructERC} replaced the ERC task from a traditional discriminative model to a generative model, and proposed a simple but effective retrieval template modulem, named InstructERC to help the model explicitly integrate multi-granular dialog supervision information.

Compared with them, DialogueLLM possesses the abilty to understand complex emotions without introducing any other components. In addition, our model would also benefit the development of task-specific LLMs.

\section{Methodology}\label{sec:model}
In this section, we detail the comprehensive pipeline for training DialogueLLM models, as shown in Fig.~\ref{fig:model}.  

\subsection{Problem Formulation}

Assume that there are $N$ conversation instances in the instruction dataset, the ${i}^{th}$ conversation $D_{i}$ contains $K$ multi-modal utterances, which is represented as $D_i= \left \{\left (C_z, M_k \right ), Y_k \right \}$, where $C_z$ denotes previous $z$ contextual utterances, $M_k$ represents the $k^{th}$ target utterance to be classified, $Y_k$ means the emotion label of the $k^{th}$ target utterance, where $i\in \left [ 1,2,...,N \right ]$, $k\in \left [ 1,2,...,K \right ]$, $z\ge 0$. The target utterance consists of textual ($T$) and visual ($V$) modalities, i.e., $M_k=\left ( T_{k}, V_{k} \right )$, where  $T_{k}\in \mathcal{R}^{l_{T_k}\times d_{T_k}}$, $V_{k}\in \mathcal{R}^{l_{I_k}\times d_{I_k}}$. Here,  $l_{T_k}$ and $l_{V_k}$ denote the sequence length of textual and visual utterances, $d_{T_k}$ and $d_{V_k}$ represents the dimensions of the textual and visual features.

Now, we summarize our research problem as: \textit{Given one multi-speaker conversation including $K$ multi-modal utterances, how to detect their emotions?} It could be written as:
\begin{equation} 
\begin{aligned}
\zeta=\prod_{k}p\left ( Y_k |C_z, M_k, \Theta  \right )
\end{aligned}
\end{equation} 
where $\Theta$ denotes the parameter set.

\subsection{Base Model}
The first key component is to select open-source and strong foundation language models. LLaMA is a collection of open source foundation language models ranging from 7B to 65B parameters, which is trained on trillions of tokens using publicly available datasets. It achieves state-of-the-art performance on numerous benchmarks, which has greatly promoted the research progress of LLMs. A considerable number of researchers choose to expand the capabilities of LLaMA models though instruction tuning, due to the lower computational costs. 

Furthermore, Meta AI has just developed and released LLaMA 2, which is an updated version of LLaMA 1. Compared with LLaMA 1, the training data used for LLaMA 2 was increased by 40\% and the context length was doubled. LLaMA 2 also incorporated grouped query attention mechanisms. LLaMA 2 shows many behaviors similar to ChatGPT, but is also surprisingly small and easy to reproduce. Hence, we adopt LLaMA 2-7B model as our base model. Furthermore, LLaMA-7B have also been attempted and evaluated in the experiments. We use low-rank adaptation (LoRA) to finetune them with only 2.1 million trainable parameters. 

In view that LLaMA 2 possess a powerful generative capability, we treat ERC as a  conditional generative task, where the output $Y_k$ will be an emotion label. We first propose a general zero-shot prompt template, namely $Prompt_{erc}^k$ that consists of the contextual, multi-modal utterances and the instruction $I_{erc}$ by merging them together:
\begin{equation} 
\begin{aligned}
Prompt_{erc}^{k}= \left \{ C_z, Text Description\left ( V_k \right ) , T_k, I_{erc} \right \} 
\end{aligned}
\end{equation} 
where $Text Description\left ( V_k \right )$ denotes the text description of the $k^{th}$ video produced by the ERNIE Bot. Then, we ask LLMs to generate an emotion label by providing the above-mentioned prompts:
\begin{equation} 
\begin{aligned}
Y_k = LLM\left ( Prompt_{erc}^{k}, \theta  \right ) 
\end{aligned}
\end{equation}

\subsection{Emotion and Context Knowledge Based Instruction Dataset}
\begin{figure}[t]
    \centering
  \includegraphics[height=1.8in, width=2.7in]{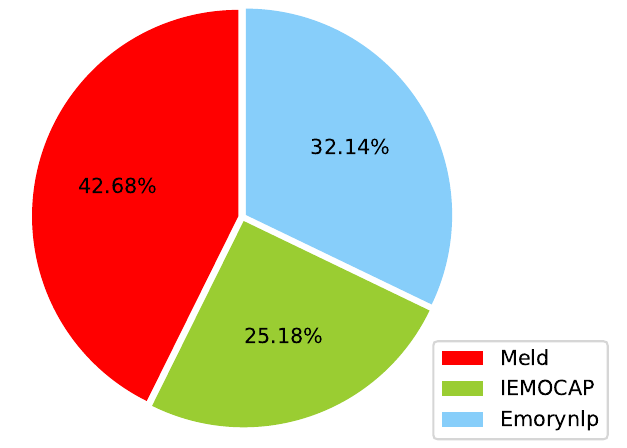}
  \caption{The distribution of three ERC datasets.}
   \label{fig:distribution1}
\end{figure}

Human conversation is filled with different emotions, such as neutral, anger, happiness, surprise, etc. To satisfy complex emotion recognition needs, DialogueLLM undergos an instruction-tuning step which involves training on supervised instruction/input/output data. Instruction tuning helps align DialogueLLM with human prompts, enabling precise customization for emotional domains. This allows DialogueLLM to become adaptable and proficient at generating accurate emotional responses. In this paper, we create a high-quality instruction dataset by leveraging three widely used benchmarking ERC datasets. Since many potential shortcomings exist in automatic generation of samples using strong language models (e.g., ChatGPT), such as low quality, repetition, and lack of diversity, etc., different from the existing works~\cite{peng2023instruction,liu2023goat}, we do not use ChatGPT to generate instances. The benchmarking ERC datasets have provided clean samples with precise annotations, which will be an optimal choice for creating instruction dataset.
The training sets of three benchmarking datasets (i.e., MELD, IEMOCAP and EmoryNLP) are treated as the data source, altogether 2,411 multi-party dialogues, over 24,304 utterances are collected. In view that the labels are from different datasets, we first pre-process the lables. For example, ``joy'', ``happy'' and ``happiness'' will be normalized to be ``happiness''. The instructions are constructed based on the task definition and label space, e.g., \textit{``Given the Video Description and Context, detect the emotion of the input, and assign an accuracy label from [`happiness', `anger', `fear', `sadness', `disgust', `surprise', `neutral'].''}. The textual raw samples and the counterpart labels are normalized to the input/output pairs. 

\begin{figure}[t]
    \centering
  \includegraphics[height=2.0in, width=3.1in]{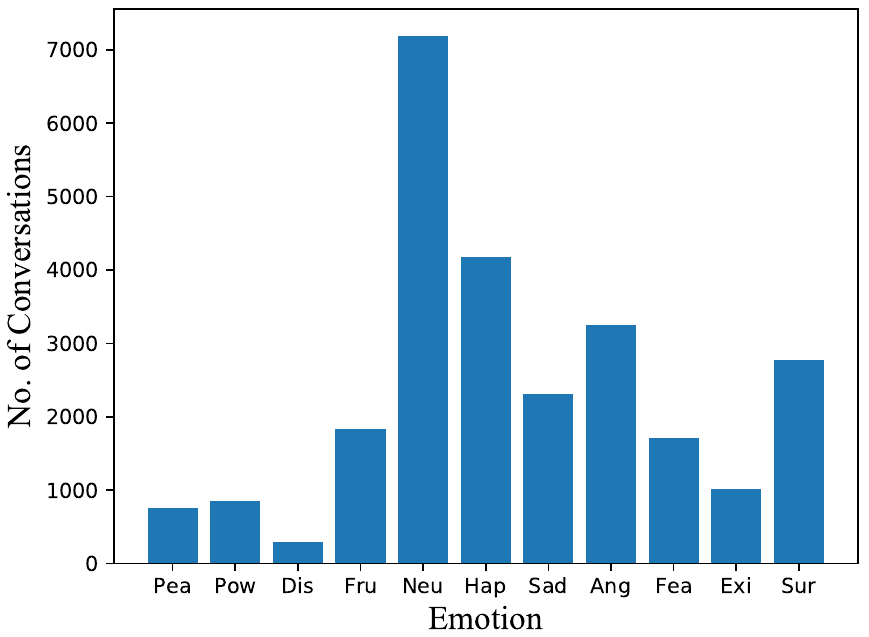}
  \caption{The distribution of seven basic emotions across three datasets.}
   \label{fig:distribution2}
\end{figure}

In view of the importance of the conversational context and multi-modal knowledge, the contextual utterances and the visual information are incorporated into instruction instances. Assume that there are $z$ contextual utterances before the target utterance, we would list them before the input content. In this work, the default size is $z=1$ (where the impact of varying size will be discussed in Sec. 4.9). In addition, the corresponding video is split into frames and forward them through the ERNIE Bot, to generate the descriptions of this video. Then, such descriptions are considered as the supplementary knowledge. More statistics of this dataset are presented in Fig.~\ref{fig:distribution1} and Fig.~\ref{fig:distribution2}. Notably, ``Neutral'' and ``Happiness'' accounted for the largest percentage of the total instances, about 31.1\% and 15.4\%, respectively. In contrast, ``Fear'', ``Powerful'', and ``Peaceful'' are represented at lower proportions. ``Fear" comprises around 6.1\% of the dataset, and ``Powerful" represents about 3.5\% of the dataset. Similarly, ``Peaceful" constitutes approximately 6.3\% of the dataset, indicating a notable but still comparatively moderate occurrence of this particular emotion. ``Anger", ``Sadness", ``Frustration", ``Surprise", ``Excitement" and ``Disgust" collectively account for around 37.6\% of the dataset. Specifically, ``Anger" accounts for about 9.7\%, ``Sadness" for about 6.4\%, ``Frustration" for about 5.6\%, ``Surprise" is about 9.4\%, ``Excitement" is about 4.2\%, and ``Disgust" is about 2.3\%. 

Finally, this instruction dataset is used for supervised fine-tuning. Notably, all the instances in the dataset are normalize as ``instruction/video descriptions/context/input/output'' pairs (see Fig.~\ref{fig:model}).

\subsection{Training and Implementation}
The DialogueLLM-7B model is fine-tuned LLaMA 2-7B with the emotional knowledge based instruction data to acquire emotion recognition skills. Training a DialogueLLM-7B model will cost about 5 hours on a 40GB A100 GPU. The total approximate tokens seen during pre-training is approximately 22 billion tokens. We optimize our model with the AdamW optimizer with the following hyper-parameters: $\beta_1 = 0.9$, $\beta_2 = 0.95$. We use a cosine learning rate schedule, such that
the final learning rate is equal to 10\% of the maximal learning rate. The activation function is set to SwiGLU to improve performance. The target utterances are forward through DialogueLLM models to generate the emotion labels. The details of
the algorithm are outlined in Algorithm~\ref{alg1}. 

\begin{algorithm}
	\renewcommand{\algorithmicrequire}{\textbf{Input:}}
	\renewcommand{\algorithmicensure}{\textbf{Output:}}
	\caption{The details of DialogueLLM}
	\label{alg1}
	\small
 \begin{algorithmic}[1]
 
	\STATE \textbf{Require: } Task description $I_{erc}$, Dataset $D$  (containing $N$ dialogues, $K$ utterances and the emotion label $Y_k$ for each utterance), Base model $LLM$
	\STATE  \textbf{Parameter: } $\theta$ 

        \STATE  \textbf{Ensure: }the emotion label $Y_{k}$

        \STATE   \textcolor{blue!60}{/* Step1: Generate \textit{Prompt} */}
        \STATE {$I_{erc}$, $D \to Prompt\left \{(I_{erc}, C_z, T_{k}, V_{k}) \right \}$ } 
        
	\STATE  \textcolor{blue!60}{/* Step 2: Input the Prompt of the $t^{th}$ sample */}
	\STATE  {$LLM(Prompt) \sim  E_{1} ,...,E_{k}$}
  
	\STATE \textcolor{blue!60}{/* Step 3: Call LoRA function for fine-tuning */}
	\STATE  {$LoRA(E_{1},...,E_{K}  )\to DialogueLLM$}
  
        \STATE \textcolor{blue!60}{/* Step 4: Generate output */}
	\STATE  $Output = DialogueLLM (I_{erc}, C_z, T_{k}, V_{k}) $
  
        \ENSURE $Y_{k}^{p}$
        
\end{algorithmic}  
\end{algorithm}

\section{Experiments}\label{sec:experiments}
\subsection{Research Question}
\textbf{RQ1:} Is it effective to propose an emotion-tailored LLM?

\textbf{RQ2:} Does modeling of the contextual dependency and multi-modal information help improve performance?

\textbf{RQ3:} Does DialogueLLM has powerful in-context learning abilities?

To answer RQ1, we compare the proposed DialogueLLM with a wide range of state-of-the-art baselines and other LLMs on three benchmark datasets in Sec.~\ref{sec:comparison}. To answer RQ2, we conduct a ablation test by removing one component at one time in Sec.~\ref{sec:ablation}. To answer RQ3, we
consider zero-shot and few-shot prompting setups, and report their results in Sec.~\ref{sec:fewshot}.

\subsection{Experimental Settings}
\textbf{Datasets.} Three benchmark ERC datasets which include the textual and visual utterances with high quality emotion annotations, are selected as the experimental beds, $viz.$ MELD\footnote{https://github.com/declare-lab/MELD.}~\cite{poria2018meld}, IEMOCAP\footnote{https://sail.usc.edu/iemocap/.}, and EmoryNLP\footnote{https://github.com/emorynlp}.

\textbf{MELD.} It consists of 13,708 multi-modal utterances from 1,433 multi-party dialogues of Friends TV series. The utterances in each dialogue are annotated with one of three sentiments (positive, negative or neutral) and one of seven emotions (anger, disgust, fear, joy, neutral, sadness or surprise). The overall Fleiss' kappa score reaches 0.43. In this work, we only use textual and visual information.

\textbf{IEMOCAP.} It is comprised of 151 recorded dialogue videos, encompassing a total of 302 videos across the entire dataset, each involving two speakers per session. The annotations for this dataset encompass 9 distinct emotions (anger, excitement, fear, sadness, surprise, frustration, happiness, disappointment, and neutrality). The recordings are distributed across five sessions, with each session featuring five pairs of speakers.

\textbf{EmoryNLP.} consists of 97 episodes, 897 scenes, and 12,606 utterances, which is a textual corpus that comprises multi-party dialogue transcripts of the Friends TV show. Each utterance is annotated with one of seven emotions, i.e., sad, mad, scared, powerful, peaceful, joyful, and neutral. The detailed statistics are shown in Table~\ref{tab:benchmark}.
\begin{table*}[h]
\small
\centering
\caption{Testing dataset statistics.}\label{tab:benchmark}
\scalebox{0.89}{
\begin{tabular}{llcccccccc}
\midrule[1pt]
\multicolumn{1}{c}{\multirow{2}{*}{\textbf{Type}}} & \multirow{2}{*}{\textbf{Dataset}} & \multicolumn{3}{c}{\textbf{Dialogue}}                & \multicolumn{3}{c}{\textbf{Utterance}}               & \multirow{2}{*}{\textbf{Class}} & \multirow{2}{*}{\textbf{Metric}} \\ \cline{3-8}\midrule[0pt]
\multicolumn{1}{c}{}                               &                                   & \textbf{Train} & \textbf{Validation} & \textbf{Test} & \textbf{Train} & \textbf{Validation} & \textbf{Test} &                                        &                                  \\ \midrule[1pt]
\multirow{3}{*}{Main Datasets}                     & MELD                              & 1,039          & 114                 & 280           & 9,989          & 1,109               & 2,610         & 7                                      & Weighted-F1                      \\
                                                   & IEMOCAP                           & \multicolumn{2}{c}{120}              & 31            & \multicolumn{2}{c}{5,810}            & 1,623         & 8                                      & Weighted-F1                      \\ 
                                                   & EmoryNLP                          & 659            & 89                  & 79            & 7,551          & 954                 & 984           & 7                                      & Weighted-F1                      \\ \midrule[1pt]

\end{tabular}
}
\end{table*}

\textbf{Evaluation metrics.} In line with the previous approaches, \textit{accuracy} (Acc) and \textit{weighted-F1} (w-F1) are used as evaluation metrics. 
For each method, we run five random seeds and report the average result of the test sets.

\textbf{Hyper-parameter.} 
We report the detailed hyperparameter settings of DialogueLLM on three datasets in Table~\ref{tab:hyper}. The maximum context length is set to 4,096. We use a weight decay of 0.1 and gradient clipping of 1.0. The batch size is set to 128. 

\begin{table}[h]
\small
\centering
\caption{Hyperparameters for fine-tuning DialogueLLM.}\label{tab:hyper}
\begin{tabular}{cc}
\midrule[1pt] 
\textbf{Hyperparameter} & \textbf{Value} \\ \midrule[1pt] 
Batch size              & 128            \\ 
Micro batch size        & 8              \\ 
Epoch                   & 10              \\ 
Learning rate           & 3e-4           \\ 
Lora r                  & 4              \\ 
Lora alpha              & 16             \\ 
Lora dropout            & 0.05           \\ 
Cutoff length           & 256            \\ \midrule[1pt] 
\end{tabular}
\end{table}

\subsection{Compared Baselines}
A wide range of SOTA baselines are included for comparison including pre-trained language model (PLM) based and LLM based approaches. They are:

\begin{itemize}

\begin{table*}[h]
\small
\centering
\caption{Comparison results  (\%) on different methods. The best scores are in bold.}\label{tab:result}
\begin{tabular}{l|c|cc|cc|cc|cc}
\midrule[1pt] 
                                                                                    &                                          & \multicolumn{2}{c|}{\textbf{MELD}}                                                                 & \multicolumn{2}{c|}{\textbf{IEMOCAP}}                                                              & \multicolumn{2}{c|}{\textbf{EmoryNLP}}                                                             & \multicolumn{2}{c}{\textbf{Avgerage}}                                                             \\ \cline{3-10} 
\multirow{-2}{*}{\textbf{Methods}}                                                  & \multirow{-2}{*}{\textbf{\# Param.}} & \multicolumn{1}{c}{Acc}                          & w-F1                         & \multicolumn{1}{c}{Acc}                          & w-F1                        & \multicolumn{1}{c}{Acc}                          & w-F1                         & \multicolumn{1}{c}{Acc}                          &w-F1                        \\ \midrule[1pt]

bc-LSTM                                                                             & 1.2M                                     & \multicolumn{1}{c|}{65.87}                                 & 64.87                                 & \multicolumn{1}{c|}{63.08}                                 & 62.84                                 & \multicolumn{1}{c|}{40.85}                                 & 36.84                                 & \multicolumn{1}{c|}{56.60}                                 & 54.85                                 \\ 

ICON                                                                                & 0.5M                                     & \multicolumn{1}{c|}{-}                                     & -                                     & \multicolumn{1}{c|}{64.00}                                 & 63.50                                 & \multicolumn{1}{c|}{-}                                     & -                                     & \multicolumn{1}{c|}{-}                                     & -                                     \\ 

MTL                                                                                 & 1.2M                                     & \multicolumn{1}{c|}{62.45}                                 & 61.90                                 & \multicolumn{1}{c|}{-}                                     & -                                     & \multicolumn{1}{c|}{36.36}                                 & 35.92                                 & \multicolumn{1}{c|}{49.40}                                 & 48.91                                 \\ 

DialogXL                                                                            & 510M                                     & \multicolumn{1}{c|}{-}                                     & 62.41                                 & \multicolumn{1}{c|}{-}                                     & 65.94                                 & \multicolumn{1}{c|}{-}                                     & 34.73                                 & \multicolumn{1}{c|}{-}                                     & 54.36                                 \\ 

TODKAT                                                                              & 330M                                     & \multicolumn{1}{c|}{67.24}                                 & 65.47                                 & \multicolumn{1}{c|}{61.11}                                 & 61.33                                 & \multicolumn{1}{c|}{42.38}                                 & 38.69                                 & \multicolumn{1}{c|}{56.91}                                 & 55.16                                 \\ 

CoG-BART                                                                            & 415.1M                                   & \multicolumn{1}{c|}{64.95}                                 & 63.82                                 & \multicolumn{1}{c|}{65.02}                                 & 64.87                                 & \multicolumn{1}{c|}{40.94}                                 & 37.33                                 & \multicolumn{1}{c|}{56.97}                                 & 55.34                                 \\ 

DialogueRNN                                                                         & 9.9M                                     & \multicolumn{1}{c|}{65.96}                                 & 65.30                                 & \multicolumn{1}{c|}{64.85}                                 & 64.65                                 & \multicolumn{1}{c|}{43.66}                                 & 37.54                                 & \multicolumn{1}{c|}{58.16}                                 & 55.83                                 \\ 

DialogueGCN                                                                         & 2.1M                                     & \multicolumn{1}{c|}{63.62}                                 & 62.68                                 & \multicolumn{1}{c|}{62.49}                                 & 62.11                                 & \multicolumn{1}{c|}{36.87}                                 & 36.43                                 & \multicolumn{1}{c|}{54.33}                                 & 53.14                                 \\ 

DialogueCRN                                                                         & 3.3M                                     & \multicolumn{1}{c|}{66.93}                                 & 65.77                                 & \multicolumn{1}{c|}{67.39}                                 & 67.53                                 & \multicolumn{1}{c|}{41.04}                                 & 38.79                                 & \multicolumn{1}{c|}{58.45}                                 & 57.36                                 \\ 

RGAT                                                                                & 13M                                      & \multicolumn{1}{c|}{-}                                     & 60.91                                 & \multicolumn{1}{c|}{-}                                     & 65.22                                 & \multicolumn{1}{c|}{-}                                     & 34.42                                 & \multicolumn{1}{c|}{-}                                     & 53.52                                 \\ 

DAG-ERC                                                                             & 9.5M                                     & \multicolumn{1}{c|}{63.75}                                 & 63.36                                 & \multicolumn{1}{c|}{66.54}                                 & 66.53                                 & \multicolumn{1}{c|}{39.64}                                 & 38.29                                 & \multicolumn{1}{c|}{56.64}                                 & 56.06                                 \\ 

CauAIN                                                                              & 6.1M                                     & \multicolumn{1}{c|}{65.85}                                 & 64.89                                 & \multicolumn{1}{c|}{65.08}                                 & 65.01                                 & \multicolumn{1}{c|}{43.13}                                 & 37.87                                 & \multicolumn{1}{c|}{58.02}                                 & 55.92                                 \\ 

COIN                                                                                & 0.6M                                     & \multicolumn{1}{c|}{-}                                     & -                                     & \multicolumn{1}{c|}{66.05}                                 & 65.37                                 & \multicolumn{1}{c|}{-}                                     & -                                     & \multicolumn{1}{c|}{-}                                     & -                                     \\ 

GraphCFC                                                                                & 0.6M                                     & \multicolumn{1}{c|}{-}                                     & 58.86                                     & \multicolumn{1}{c|}{-}                                 & 68.91                                & \multicolumn{1}{c|}{-}                                     & -                                     & \multicolumn{1}{c|}{-}                                     & -                                     \\ 

SACL-LSTM                                                                           & 2.6M                                     & \multicolumn{1}{c|}{{\color[HTML]{F8A102} \textbf{67.51}}} & {\color[HTML]{F8A102} \textbf{66.45}} & \multicolumn{1}{c|}{{\color[HTML]{F8A102} \textbf{69.08}}} & {\color[HTML]{F8A102} \textbf{69.22}} & \multicolumn{1}{c|}{{\color[HTML]{F8A102} \textbf{42.21}}} & {\color[HTML]{F8A102} \textbf{39.65}} & \multicolumn{1}{c|}{{\color[HTML]{F8A102} \textbf{59.60}}} & {\color[HTML]{F8A102} \textbf{58.44}} \\ 
\midrule[0.5pt]

LLaMA-7B                                                                               & 2.1M                                       & \multicolumn{1}{c|}{15.09}                                     & 16.02                                     & \multicolumn{1}{c|}{19.32}                                     & 18.24                                     & \multicolumn{1}{c|}{17.78}                                     & 17.40                                     & \multicolumn{1}{c|}{17.40}                                     & 17.22                                    \\

Alpaca                                                                              & 2.1M                                       & \multicolumn{1}{c|}{19.22}                                     & 18.37                                     & \multicolumn{1}{c|}{20.35}                                     & 19.16                                     & \multicolumn{1}{c|}{17.95}                                     & 17.33                                     & \multicolumn{1}{c|}{19.17}                                     & 18.29                                     \\

LLaMA 2-7B                                                                             & 4.2M                                       & \multicolumn{1}{c|}{23.71}                                     & 24.12                                     & \multicolumn{1}{c|}{26.73}                                     & 24.35                                     & \multicolumn{1}{c|}{25.50}                                     & 17.27                                     & \multicolumn{1}{c|}{25.31}                                     & 21.91                                     \\  \midrule[0.5pt]

                                                                                    & {4.2M}       & \multicolumn{1}{c|}{{\color[HTML]{FE0000} \textbf{71.96}}}     & {\color[HTML]{FE0000} \textbf{71.90}}     & \multicolumn{1}{c|}{{\color[HTML]{FE0000} \textbf{70.62}}}     & {\color[HTML]{FE0000} \textbf{69.93}}     & \multicolumn{1}{c|}{{\color[HTML]{FE0000} \textbf{41.88}}}     & {\color[HTML]{FE0000} \textbf{40.05}}     & \multicolumn{1}{c|}{{\color[HTML]{FE0000} \textbf{61.49}}}     & {\color[HTML]{FE0000} \textbf{60.52}}     \\

\multirow{-2}{*}{\begin{tabular}[c]{@{}l@{}}DialogueLLM\\ Improve $\bigtriangleup $\end{tabular}}  &       & \multicolumn{1}{c|}{ $\uparrow 6.59\%$}     & {$\uparrow 8.20\%$}     & \multicolumn{1}{c|}{ $\uparrow 2.22\%$}     & {$\uparrow 1.03\%$}     & \multicolumn{1}{c|}{$\downarrow 0.78\%$}     & {$\uparrow 1.00\%$}     & \multicolumn{1}{c|}{$\uparrow 3.17\%$}     & { $\uparrow 3.56\%$}   \\ 

\midrule[1pt] 
\end{tabular}
\end{table*}

\item \textit{\textbf{PLM based approaches:}}

  \textbf{(1) bc-LSTM~\cite{zhang2023learning}} implements an utterance-level LSTM to capture contextual features.

 \textbf{(2) ICON~\cite{hazarika2018icon}} hierarchically models the self- and inter-speaker emotional influences into global memories, and generates contextual summaries.

  \textbf{(3) MTL~\cite{Li2020MultiTaskLW}} exploits speaker identification (SI) as an auxiliary task to enhance the utterance representation in conversations.

 \textbf{(4) DialogXL~\cite{shen2021dialogxl}} modifies the recurrence mechanism of XLNet to store longer historical context and dialog-aware self-attention to deal with the multi-party structures.

 \textbf{(5) TODKAT~\cite{zhu-etal-2021-topic}} designs a transformer-based encoder-decoder architecture fuses the topical and commonsense information, and performs the emotion label sequence prediction.

 \textbf{(6) CoG-BART~\cite{li2022contrast}} uses the pre-trained encoder-decoder model BART as the backbone model and utilizes an auxiliary response generation task to enhance the model's ability of handling context information.

 \textbf{(7) DialogueRNN~\cite{majumder2019dialoguernn}} designs a method based on recurrent neural networks (RNN) that keeps track of the individual party states throughout the conversation and uses this information for emotion classification.

 \textbf{(8) DialogueGCN~\cite{ghosal2019dialoguegcn}} leverages self and inter-speaker dependency of the interlocutors to model conversational context for emotion recognition.

 \textbf{(9) DialogueCRN~\cite{hu2021dialoguecrn}} designs multi-turn reasoning modules to extract and integrate emotional clues.

 \textbf{(10) RGAT~\cite{ishiwatari2020relation}} proposes relational position encodings  to capture both the speaker dependency and the sequential information.

 \textbf{(11) DAG-ERC~\cite{ghosal2019dialoguegcn}}  regards each conversation as a directed acyclic graph to model the conversation context.

 \textbf{(12) CauAIN~\cite{zhao2022cauain}} retrieves causal clues provided by commonsense knowledge to guide the process of causal utterance traceback.

 \textbf{(13) COIN~\cite{zhang2021coin}}  is a conversational interactive model to mitigate the problem of  overlooking the immediate mutual interaction between different speakers by applying state mutual interaction within history contexts.
 
 \textbf{(14) GraphCFC~\cite{DBLP:journals/corr/abs-2207-12261}} is a module adept at modeling context and interaction information in ERC tasks with high efficiency. It leverages multiple extractors and PairCC strategies to effectively tackle the heterogeneity present in multimodal fusion.

 \textbf{(15) SACL-LSTM~\cite{hu2023supervised}}  applies contrast-aware adversarial training to generate worst-case samples and uses a joint class-spread contrastive learning objective on both original and adversarial samples.

\item \textit{\textbf{LLMs based approaches:}}

  \textbf{(1) LLaMA~\cite{touvron2023llama}} takes a sequence of words as an input and predicts a next word to recursively generate text.

  \textbf{(2) Alpaca} is a state-of-the-art finedtuning version of LLaMA, by using supervised learning from a LLaMA 7B model on 52K instruction-following demonstrations.

  \textbf{(3) LLaMA 2~\cite{touvron2023llama2}} is trained on 2 trillion tokens, and have double the context length than Llama 1, and outperforms other open source language models on many external benchmarks.

\end{itemize}

\subsection{Results and Anlysis}\label{sec:comparison}
The experimental performance of all baselines is shown in Table~\ref{tab:result}. We divide these baselines into two categories, i.e., pre-trained language models and large language models without fine-tuning. We will conduct a detailed analysis of their classification performance.

For the PLM based baselines, we can observe that MTL demonstrates very poor performance compared to the other baseline models, with the worst classification accuracy on the MELD and EmoryNLP datasets. One possible reason is that MTL disregards modeling conversation-level interaction information. Without capturing contextual information, the model struggles to learn effectively, resulting in inaccurate classification outcomes. In contrast, DialogueCNN, DialogueGCN, and DialogueCRN aim to model the contextual information derived from the speaker. Compared to MTL, their performance substantially improves. This further verifies that incorporating contextual information is vital for ERC.

The SACL-LSTM model performs very well, being the second-best in terms of the average scores on three datasets. It may benefit from an architecture that combines the strengths of LSTM with self-attention mechanisms, enabling it to capture both long-term dependencies and subtle contextual cues within dialogues. Other strong models like TODKAT, CoG-BART, and DialogXL show competitive performance but do not reach the same level as SACL-LSTM across MELD and IEMOCAP datasets. They do not achieve top scores, which could be attributed to the complexity of emotion recognition tasks that may not be entirely captured by the models' pre-training data or architecture. With 6.1M parameters, CauAIN demonstrates solid performance, especially on the MELD and IEMOCAP datasets. This model's architecture likely includes mechanisms that aid in capturing causal relationships within dialogues, which is a critical factor for understanding emotions. The COIN model, despite its smaller size of 0.6M parameters, still achieves competitive accuracy on the IEMOCAP dataset. But it was not evaluated on other datasets.

In addition, Table~\ref{tab:result} shows that when LLMs are fine-tuned without using the proposed emotional knowledge, all of LLaMA-7B, Alpaca, and LLaMA2-7B perform very poor on three emotion recognition tasks. This proves that the general priori knowledge of LLMs are not sufficient to handle complex and subjective emotion understanding tasks. The emotion-specific knowledge is needed to further deepen their potential. In contrast, The proposed DialogueLLM model achieves the state-of-the-art performance across three datasets, which proves the effectiveness of fine-tuning LLMs with task specific knowledge. 

\textbf{MELD.} DialogueLLM achieves remarkable results on the MELD dataset, demonstrating its robustness with a leading F1 score of 71.90\% and an accuracy of 71.96\%. This is a significant uptick from the other strong contender on this dataset, SACL-LSTM, which registers an F1 score of 66.45\% and an accuracy of 67.51\%. The MELD dataset, known for its realistic conversational scenarios from a popular TV show, poses a challenging benchmark due to its diverse emotional expressions and informal dialogue. DialogueLLM's performance here suggests its superior ability to decode nuanced emotional cues within a naturalistic dialogue setting.

\textbf{IEMOCAP.} DialogueLLM showcases its prowess with an accuracy of 70.62\% and an F1 score of 69.93\%, outstripping the previously leading SACL-LSTM model, which had an accuracy of 69.08\% and an F1 score of 69.22\%. The IEMOCAP dataset is unique due to its focus on dyadic conversations with a rich set of emotional annotations, ranging from anger to happiness. The high performance of DialogueLLM on this dataset underscores its effectiveness in understanding and interpreting complex emotional dynamics in close-ended conversations.

\textbf{Emorynlp.} DialogueLLM maintains a competitive edge, securing an F1 score of 40.05\% and an accuracy of 41.88\%. DialogueLLM consistently surpasses all of the 15 baselines. This performance is indicative of DialogueLLM's versatile capacity to capture emotional nuances across varied conversational contexts

The experimental results demonstrate the effectiveness of the DialogueLLM model in emotion recognition tasks across different datasets. It consistently achieves the highest F1 scores and accuracy, outperforming 15 state-of-the-art models. Notably, DiaologueLLM's performance is robust across various datasets, which underscores its versatility and reliability in handling different data sources and domains. This also suggests a trend where specialized pre-training on tasks closely related to the downstream application can yield significant benefits.

\begin{figure*}[t]
    \centering
  \includegraphics[height=1.9in, width=6in]{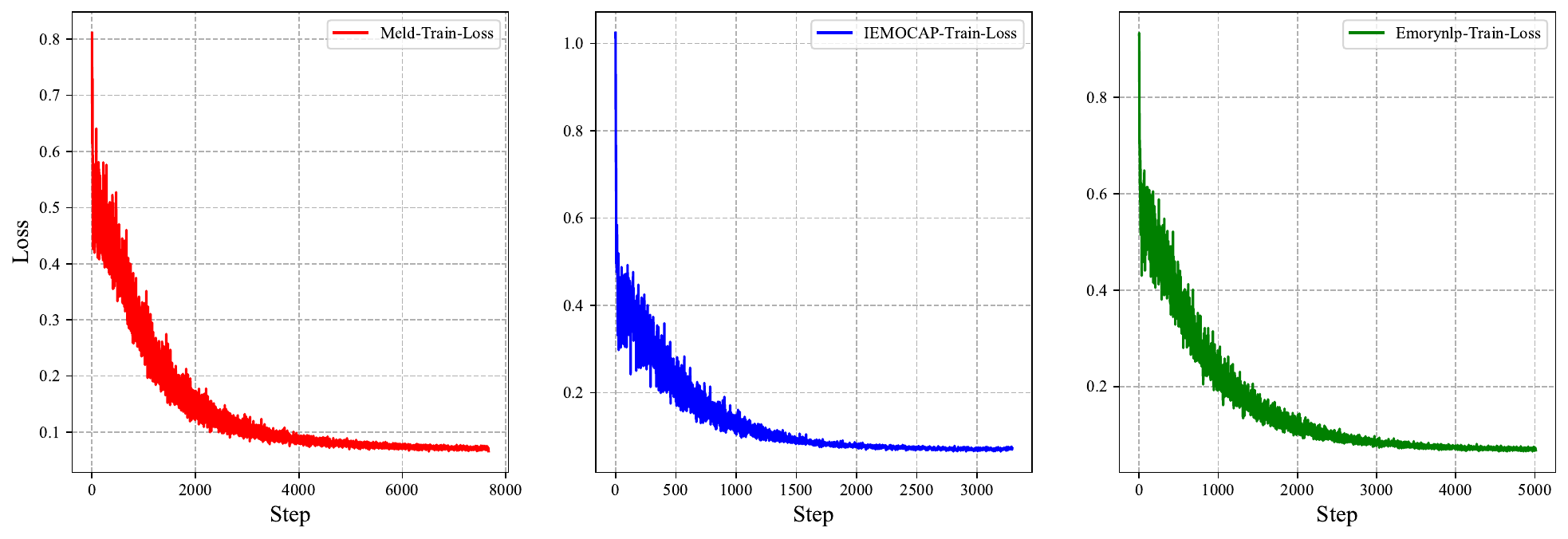}
  \caption{The training loss of DialogueLLM.}
   \label{fig:loss}
\end{figure*}

\textit{Training loss.} The training loss is shown in Figure~\ref{fig:loss}. The loss result on MELD dataset shows a rapid initial decrease with some early fluctuations, eventually stabilizing at around a 0.2 loss value. The IEMOCAP dataset's training loss drops more abruptly than MELD, suggesting a faster learning rate or easier dataset for the model to learn, and levels off at a lower value near 0.1, indicating a more successful training outcome. Lastly, the EmoryNLP dataset's loss decreases smoothly without the volatility observed in the MELD graph, also stabilizing at a loss value just under 0.2. This smooth decrease may point to a stable learning process. 

\subsection{Ablation Test}\label{sec:ablation}
The ablation study is conducted across three datasets, which will provide a structured insight into the contribution of different components to the model's performance. The term ``w/o'' indicates the model's performance without a specific feature, where the term ``w'' indicates the model's performance with a specific feature.

From Table~\ref{tab:ablation}, we have four observations: (1) the DialogueLLM's performance decreases across all datasets when any component is removed, underscoring the integral role each part plays in the model's design for emotion recognition; (2) the performance drops with the removal of context on all datasets suggests that contextual information is important for emotion recognition, aligning with the premise that conversational emotion understanding is heavily reliant on context; (3) the removal of LoRA leads to a significant decrease in model performance, because the small training size leads to underfitting; (4) removing the visual information leads to a noticeable decrease in performance, suggesting that multimodal information may be beneficial for emotion recognition in dialogues. Notably, we do not use the visual information from IEMOCAP, because the actors in this corpus are sitting on chairs for face-to-face conversations, and the descriptions of the image information are too similar, e.g., ``a man and a woman sitting on chairs for face-to-face exchanges''. Here, we have given the answer to RQ2.

\begin{table}[h]
\small
\begin{center}
\caption{\label{tab:ablation} Ablation experiment results across three ERC tasks in a zero-shot setting.}
\begin{tabular}{clcc}
\midrule[1pt]    
\multirow{1}*{\textbf{Dataset}} & \multirow{1}{0.2\textwidth}{\centering\textbf{Models}}   & \textbf{Acc}  & \textbf{w-F1}\\

\midrule[1pt]  
\multirow{4}*{MELD} & \textit{w/o} Context & 70.91 & 67.94  \\ 
& \textit{w/o} Lora & 66.17 & 64.42  \\ 
 & \textit{w/o} Video Description & 60.80 & 59.75      \\ 
\cline{2-4}
\midrule[0pt] 
& DialogueLLM & \textbf{71.91} & \textbf{71.81}     \\
\midrule[1pt]

\multirow{4}*{IEMOCAP} &  \textit{w/o} Context & 68.14 & 68.01    \\ 
& \textit{w/o} Lora & 65.23 & 63.78  \\ 
 & \textit{w/o} Video Description & - & -      \\ 
\cline{2-4}
\midrule[0pt] 
& DialogueLLM & \textbf{70.48} & \textbf{69.40}     \\
\midrule[1pt]

\multirow{4}*{Emorynlp} &  \textit{w/o} Context & 39.41 & 36.25   \\ 
& \textit{w/o} Lora & 35.66 & 33.83  \\ 
 & \textit{w/o} Video Description & - & -      \\ 
\cline{2-4}
\midrule[0pt] 
& DialogueLLM & \textbf{41.76} & \textbf{38.47}      \\

\midrule[1pt]    
\end{tabular}
\end{center}
\end{table}

\subsection{Zero-shot v/s Few-shot Prompting}\label{sec:fewshot}
This paper also performs zero-shot and few-shot experiments to evaluate whether DialogueLLM can perform better when a limited number of cases are available for emotion recognition tasks. The results are shown in Table~\ref{sec:Few-shot}. We design four $H$-shot settings: zero-shot, one-shot, five-shot, ten-shot. For each setting, we sample $H= \left \{ 0,1,5,10 \right \} $ examples for emotion classification. These sampling examples serve as the learning samples for DialogueLLM. 

The impact of adding shots varies with the number of shots. The performance gains are not significant or even decreased when adding too many shots. The change from zero-shot to one-shot results in a slight improvement in classification performance. With the gradual increase in the number of shots, the performance drops down.
 
This could be attributed to misclassifications made by DialogueLLM, potentially arising from the model learning excessive redundant information when handling too long contextual data. This suggests that roughly increasing the number of extra shots does not necessaryly result in a stable performance improvement.

\begin{table}[h]
\centering
\caption{Few shot performance of emotion recognition task.}\label{sec:Few-shot}
\scalebox{0.8}{
\begin{tabular}{c|cc|cc|cc}
\toprule
\multirow{2}{*}{\textbf{Prompt}} & \multicolumn{2}{c|}{\textbf{Meld}}      & \multicolumn{2}{c|}{\textbf{IEMOCAP}}   & \multicolumn{2}{c}{\textbf{Emorynlp}} \\ \cline{2-7} 
                        & Acc & \multicolumn{1}{l|}{w-F1} & Acc & \multicolumn{1}{l|}{w-F1} & Acc & \multicolumn{1}{l}{w-F1} 

                        \\  \midrule[1pt] 
0-shot                  &71.91   &{71.81}   &{70.48}     & {69.40}  & {41.76}    &{38.47}           \\ 
1-shot                  &71.96     &71.90                         & 70.62    & 69.93                        & 41.88              &40.05            \\ 
5-shot                  &71.92     &71.63                        &70.54     & 69.37                        &41.62               &38.61           \\ 
10-shot                 &70.65     &71.04                         &69.94     &68.82                         &41.25               &38.19            \\ \midrule[1pt] 
\end{tabular}}

\end{table}

\subsection{Error Analysis}\label{sec:error}

\begin{figure}[h]
    \centering
  \includegraphics[height=2.0in, width=2.3in]{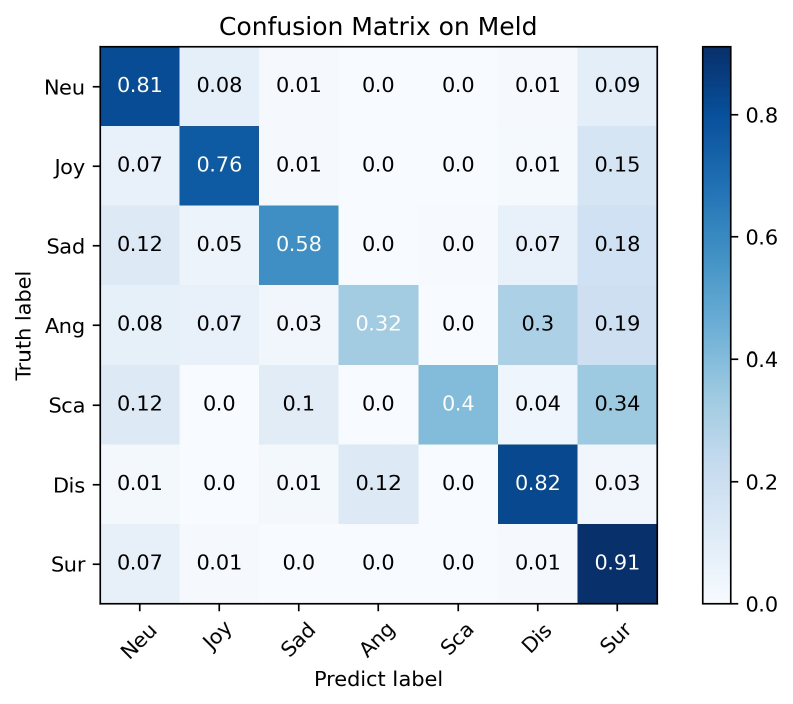}
  \includegraphics[height=2.0in, width=2.4in]{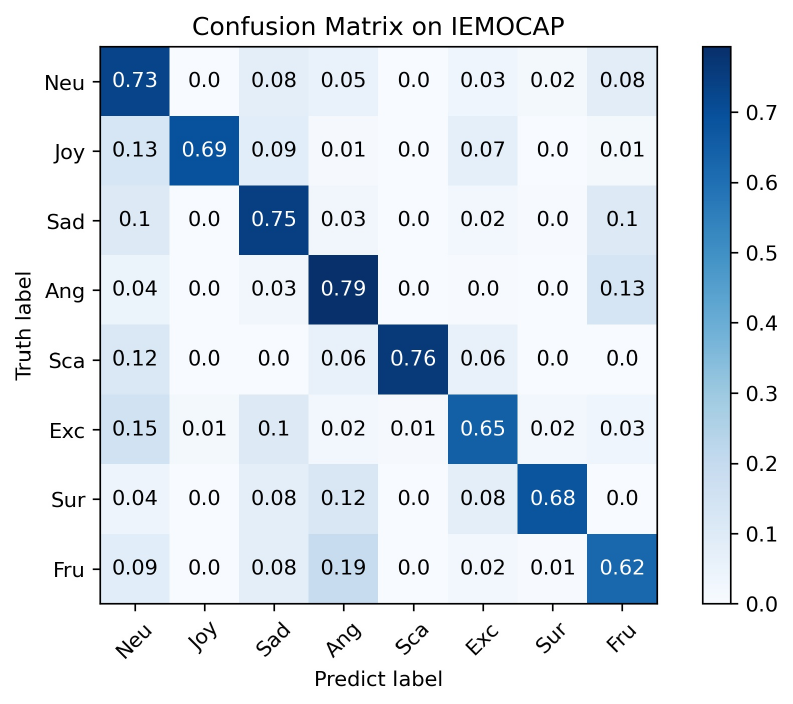}
  \includegraphics[height=1.95in, width=2.3in]{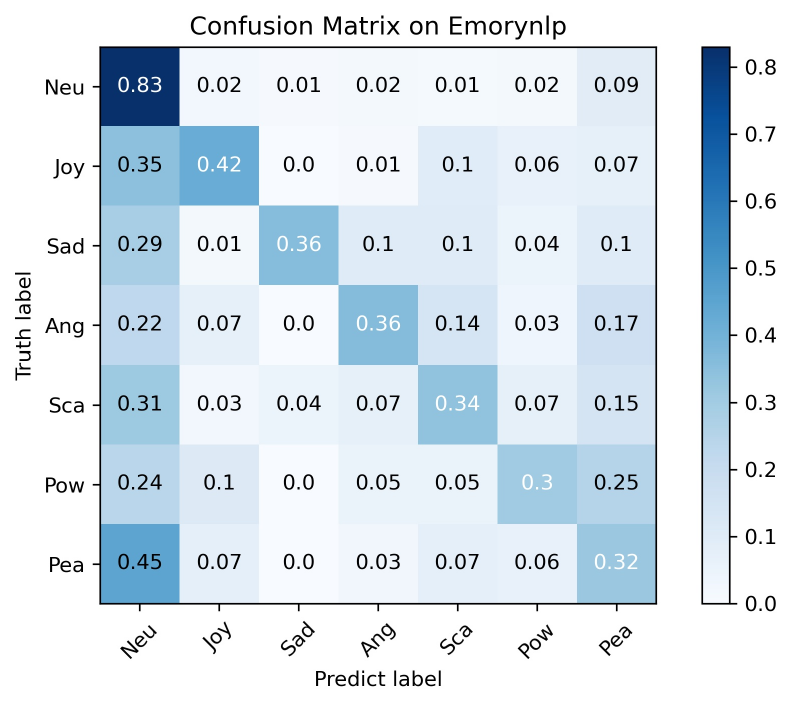}
  \caption{ The normalized confusion matrices for DialogueLLM. The rows represent the truth label, where the columns represent the predicted labels. }
   \label{fig:confusion matrix}
\end{figure}

The detailed error analysis is also conducted via the confusion matrices that are shown in Figure~\ref{fig:confusion matrix}.

Each cell $\left ( i,j \right ) $ represents the percentage of class $i$ is classified to be class $j$. Upon reviewing the classification results produced by DialogueLLM on the three datasets, we discover that imbalanced emotion categories and the similarity across different emotions are the key factors contributing to misclassification.

By examining the diagonal elements of the matrices, DialogueLLM demonstrates effective true-positive categorization for most fine-grained emotions. However, it exhibits a tendency to misclassify the utterances to be ``neutral'', particularly in the EmoryNLP dataset. This misclassification is influenced by the dataset's imbalance, where the percentages of ``neutral'' utterances in EmoryNLP, MELD and IEMOCAP are 32.48\%, 47.5\%, and 22.23\%, respectively. This highly unbalanced data distribution leads to the model's excessive preference for ``neutral'' emotions.

\begin{figure*}[h]
    \centering
  \includegraphics[height=3.5in, width=6.4in]{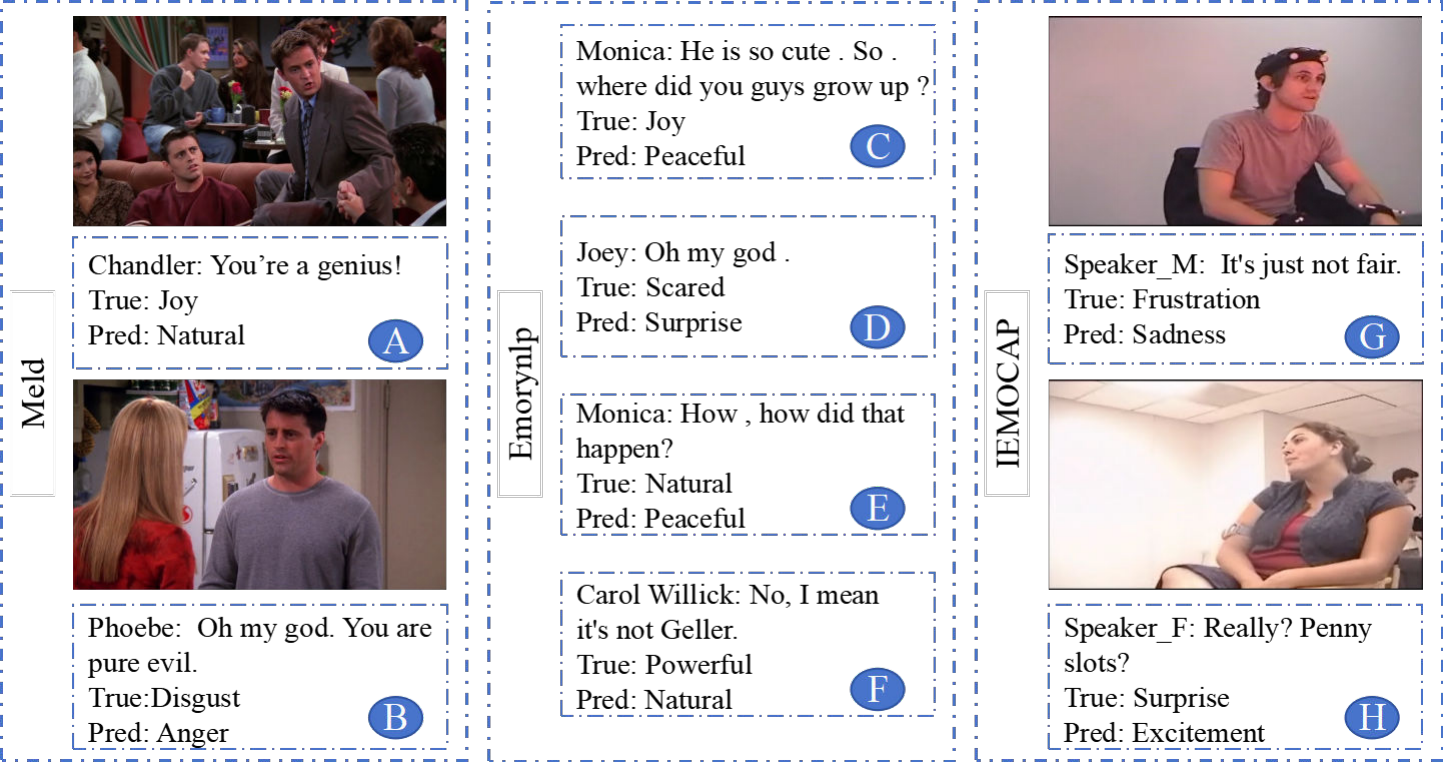}
  \caption{A few typical misclassified examples on three datasets. }
   \label{fig:error}
\end{figure*}

Additional, we show a few typical misclassification examples in Figure~\ref{fig:error}. It is evident that DialogueLLM encounters challenges in distinguishing closely related pairs of emotions. In the confusion matrices, we observe a consistent misclassification of ``anger''  to be ``disgust'' on the MELD dataset, see the example $B$ in Figure~\ref{fig:error}. In this case, when Phoebe makes a negative comment to another person, it is challenging to discover whether the expressed emotion is ``disgust'' or ``anger''. A few pairs of emotions such as ``surprise'' vs ``excitement'', ``anger'' vs ``frustration'' and ``peaceful'' vs ``happiness''. The slight similarity across such emotions poses a challenge for the model to accurately distinguish them. This difficulty in discerning emotions may result in errors during emotion categorization.

\subsection{The Impact of Epoch}\label{sec:epoch}

\begin{figure*}[h]
    \centering
  \includegraphics[height=2.1in, width=6.1in]{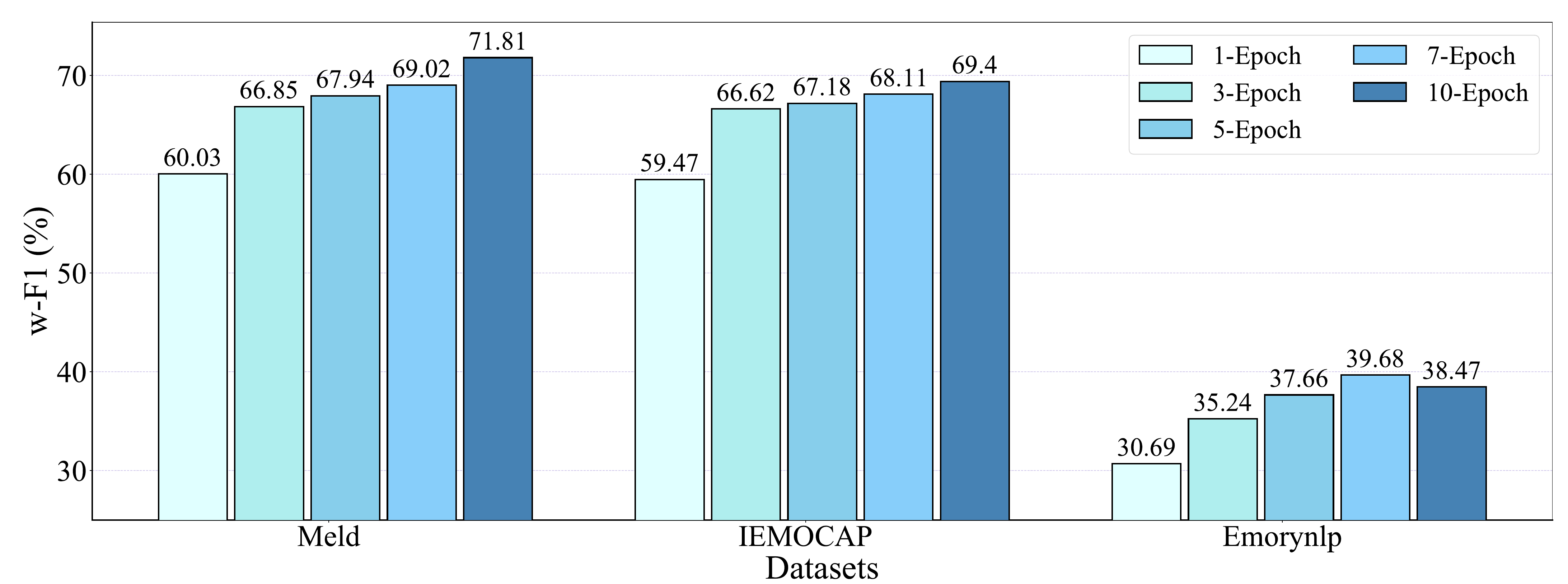}
  \caption{ The impact of epoch on the classification performance in a zero-shot setting.  }
   \label{fig:epoch1}
\end{figure*}

In this section, the impact of the number of epoch on the classification performance is shown in Figure~\ref{fig:epoch1}. We can notice that the performance increases with iterative epoch on all three datasets. 

In particular, there is a sharp leap in performance when epoch ranges from 1 to 3, and a slow increase in performance when epoch ranges from 3 to 10. However, the performance slightly decreases when epoch varies from 7 to 10 on the EmoryNLP dataset. One possible explanation is overfitting on the training set due to the small size of this dataset. This result shows the importance of selecting an appropriate number of iteration rounds. Adding the number of epoch often leads to longer training times. Due to the limitation of our GPUs, we refrained from testing with a higher number of epochs.

\begin{figure*}[h]
    \centering
  \includegraphics[height=2in, width=5.3in]{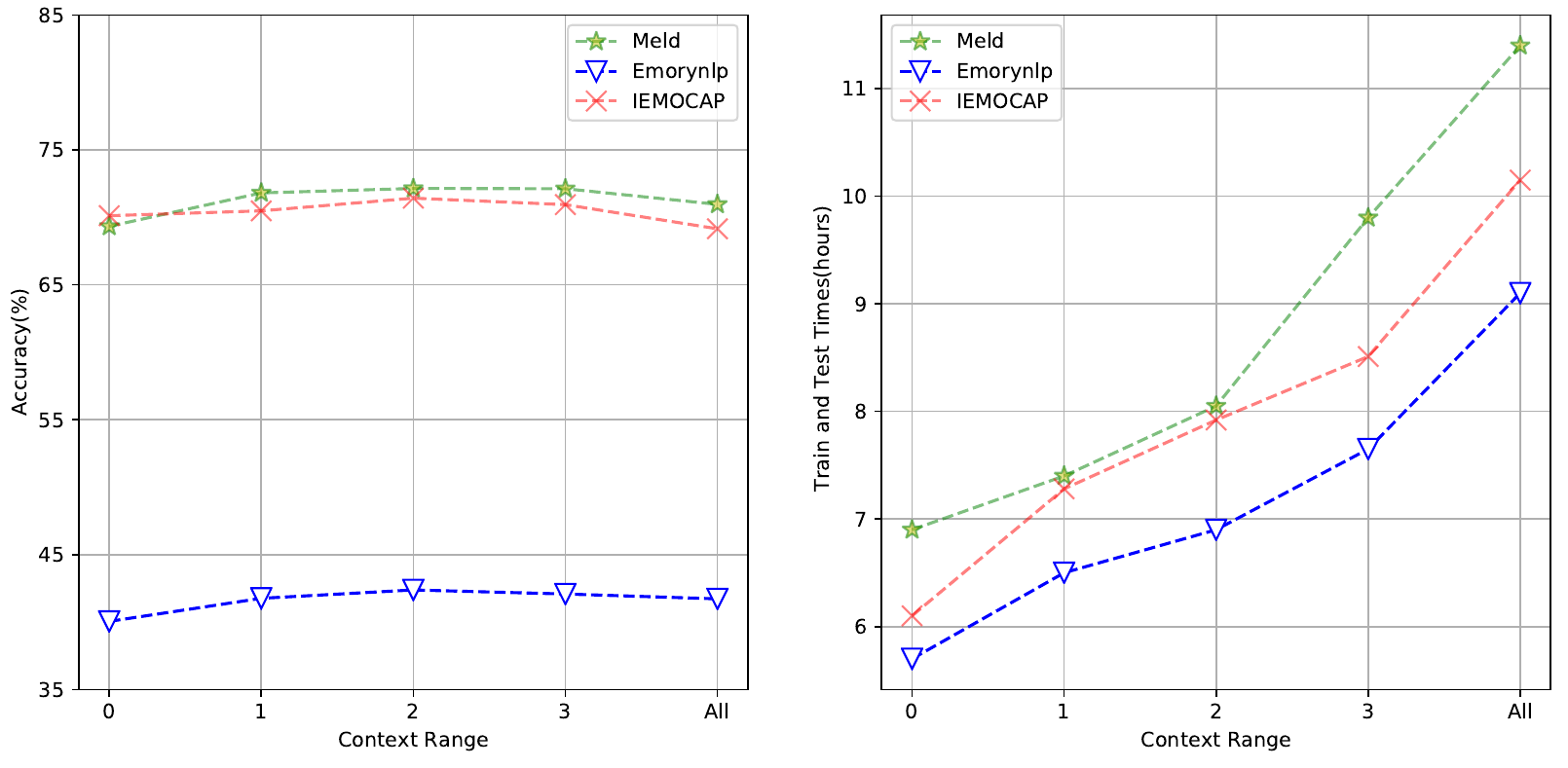}
  \caption{The experiments of the varying context lengths.}
   \label{fig:context}
\end{figure*}

\subsection{Effect of Context Range}\label{sec:context}
 
This subsection aims to explore the impact of context length on the performance of DialogueLLM. We select and evaluate the context length from the pool $\left \{ 0,1,2,3,All \right \} $, where \textit{All} represents using all the contexts. The experimental results are shown in Figure~\ref{fig:context}.

In general, longer context length will allow the model to access more information, thus making accurate prediction. We can notice that there is a slight improvement when the context length ranges from 0 to 2, and the performance slightly decreases when the context length ranges from 3 to All. This shows that too short can not provide supplementary knowledge where too long will introduce excessive amounts of noise. Hence, taking the previous two utterances before the target utterance into consideration may be a optimal choice. Additionally, processing long contextual information demands increased computational resources, thereby constraining the model's utility.
 
 \begin{figure}[h]
    \centering
  \includegraphics[height=2.7in, width=3.0in]{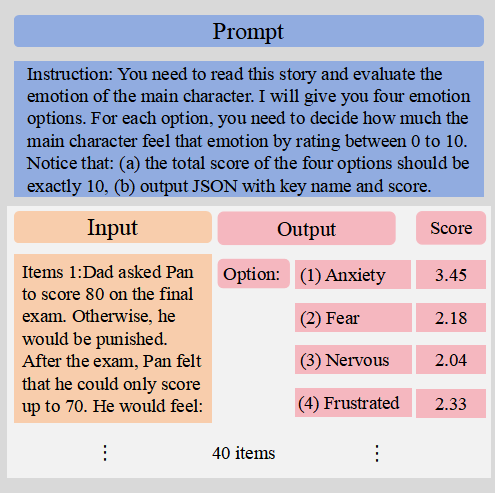}
  \caption{ The proposed prompt for the emotional intelligence test\footnote{https://emotional-intelligence.github.io/}.}
   \label{fig:EQ}
\end{figure}

\subsection{Analysis of DialogueLLM's Emotional Intelligence}\label{sec:EQ}
Emotional intelligence (EQ) is the ability to manage both human emotions and understand the emotions of other people. A people's EQ affects his/her daily behavior and decision making. Since LLMs (including DialogueLLM) have shown strong emotion understanding ability, then evaluating their emotion intelligence will be the new target. We will answer the question: \textit{can DialogueLLM be as emotionally intelligent as humans?}

In particular, we evaluate 12 LLMs' EQ via a benchmarking testing bed, namely SECEU~\cite{EQ}. This test requires evaluating complex emotions (e.g., surprised, joyful, puzzled, proud) in realistic scenarios (e.g., despite feeling underperformed, John surprisingly achieved a top score). According SECEU's requirement, we design a standard prompt to ask LLMs to do multi-choice questions in SECEU, where the prompt template is illustrated in Figure~\ref{fig:EQ}.

The model is tasked with scoring the extent to which the protagonist experiences a given emotion on a scale from 0 to 10, with 10 indicating the highest possible level of that emotion. The cumulative score for the last four emotions must be constrained to 10. To standardize the performance of LLMs, we calculate the Euclidean distance between the LLMs' responses for the $i^{th}$ item (denoted as $L_i$) and the standard human scores (denoted as $SS_i$). We then average all 40 distances and generate the SECEU score. We then normalize the SECEU scores to derive an EQ score designed to conform to a normal distribution with a mean of 100 and a standard deviation of 15. The calculating processing of EQ is written as:
\begin{equation} 
\begin{aligned}
LS_{i} = \sum_{k=1}^{4} (L_{i1}-SS_{i1}  )^{2}
\end{aligned}
\end{equation} 

\begin{equation} 
\begin{aligned}
SECEU_{score} = \frac{1}{40} \sum_{i=0}^{39} \sqrt{LS_{i} } 
\end{aligned}
\end{equation}

\begin{equation} 
\begin{aligned}
EQ = 15 \times \frac{M-SECEU_{score}}{SD} + 100
\end{aligned}
\end{equation}
where $M=2.79$ and $SD=15$ represents the mean value and the standard deviation.

\begin{figure}[h]
    \centering
  \includegraphics[height=2.0in, width=3.05in]{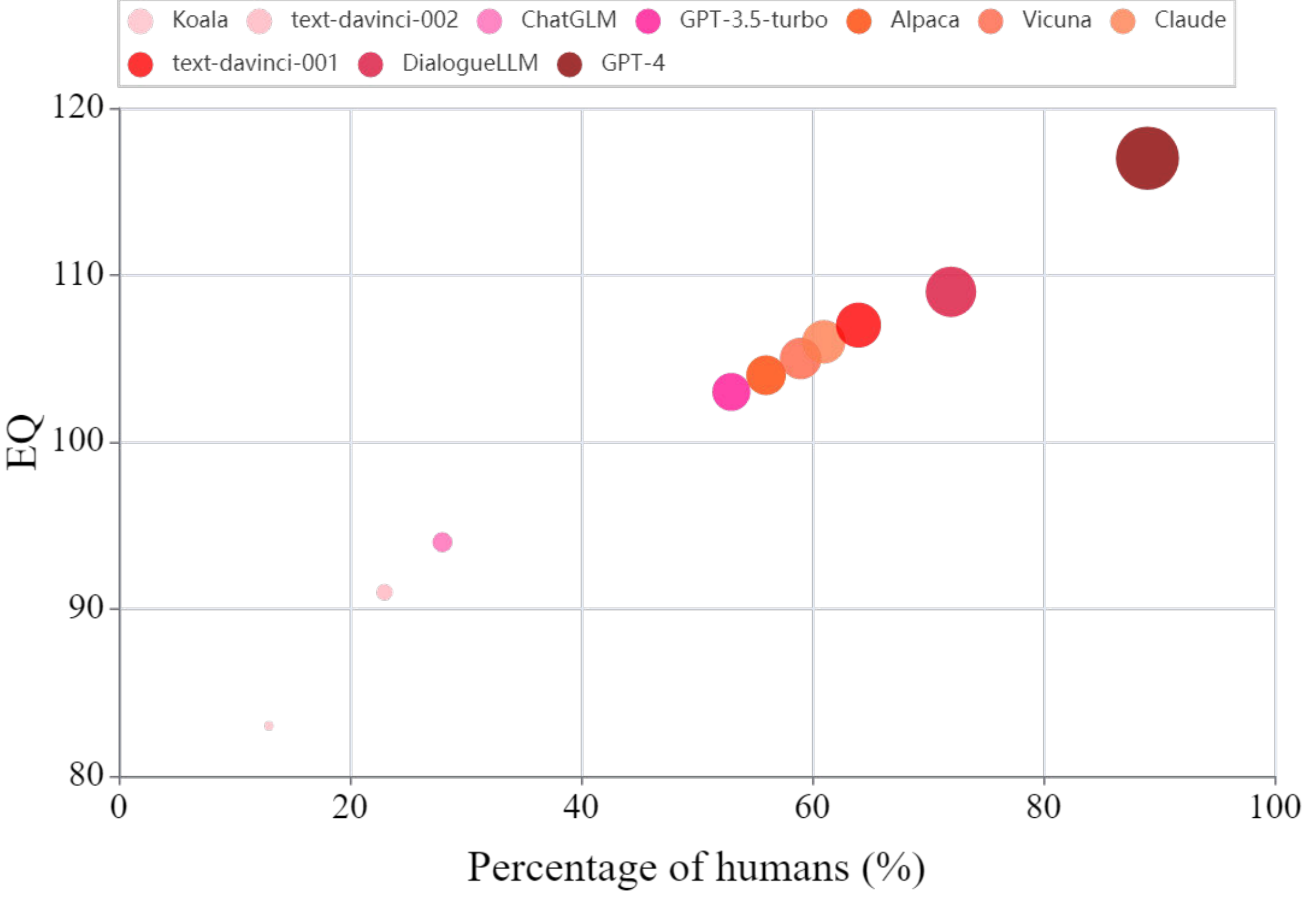}
  \caption{LLMs' EQ. The y-axis indicates the EQ score and the x-axis shows the percentage of total participants.}
   \label{fig:eqfig}
\end{figure}

The results of 12 LLMs' EQ scores are shown in Table 6 and Figure~\ref{fig:eqfig}. We can notice that the LLaMA base model cannot complete the EQ test, where our DialogueLLM achieves the second highest scores across 12 LLMs. GPT-4 achieves the highest EQ scores against ours (117 \textit{v/s} 109). However, our DialogueLLM model has only 7 billion parameters, which is 1/257 of GPT-4 (1.8 trillion parameters). The training time required for DialogueLLM is considerably shorter than that of GPT-4.
DialogueLLM performs the best among all the LLaMA series, and  exhibits human-like response patterns, demonstrating a balanced mechanism for high emotion understanding proficiency.

\begin{table}[h]
\small
\begin{center}
\caption{\label{tab:EQscore} LLMs' EQ results. \% indicates the percentage of LLM who outperforms humans in this EQ test.}
\scalebox{0.85}{
\begin{tabular}{llcccc}
\midrule[1pt] 
\textbf{Based} & \multicolumn{1}{c}{\textbf{Models}}  & \textbf{SECEU}       & \textbf{EQ}          & \textbf{\%}        & \textbf{Size} \\ \midrule[1pt] 
\multicolumn{2}{l}{\textbf{OpenAI GPT series}}        & \multicolumn{1}{l}{} & \multicolumn{1}{l}{} & \multicolumn{1}{l}{} \\
               & \multicolumn{1}{c}{text-davinci-001} & 2.4                  & 107  & 64\%                & \textless{}175B      \\
               & \multicolumn{1}{c}{text-davinci-002} & 3.3                  & 91   & 23\%               & \textless{}175B      \\
               & GPT-3.5-turbo                        & 2.63                 & 103   &52\%               & 175B                 \\
               & GPT-4                                & 1.89                 & 117   &89\%               & 1800B \\ \midrule[0.5pt] 
\multicolumn{2}{l}{\textbf{LLaMA}}                    & \multicolumn{1}{l}{} & \multicolumn{1}{l}{} & \multicolumn{1}{l}{} \\
               & LLaMA                                &Failed                   & --  &--                 & 7B                   \\
               & Alpaca                               & 2.56                 & 104  &56\%                & 13B                  \\
               & Koala                                & 3.72                 & 83   &13\%                & 13B                  \\
               & Vicuna                               & 2.5                  & 105  &59\%                & 13B                  \\ \midrule[0.5pt] 
\multicolumn{2}{l}{\textbf{Flan-t5}}                  & \multicolumn{1}{l}{} & \multicolumn{1}{l}{} & \multicolumn{1}{l}{} \\
               & Fastchat                             & Failed                   & --  &--                 & 3B                   \\ \midrule[0.5pt] 
\multicolumn{2}{l}{\textbf{GLM}}                      & \multicolumn{1}{l}{} & \multicolumn{1}{l}{} & \multicolumn{1}{l}{} \\
               & ChatGLM                              & 3.12                 & 94   &28\%                & 6B                   \\ \midrule[0.5pt] 
 
 \multicolumn{2}{l}{\textbf{Claude}}                      & \multicolumn{1}{l}{} & \multicolumn{1}{l}{} & \multicolumn{1}{l}{} \\
               & Claude                              & 2.46                 & 106   &61\%                & 10B                   \\ \midrule[0.5pt]            
               
\multicolumn{2}{l}{\textbf{Ours}}                     & \multicolumn{1}{l}{} & \multicolumn{1}{l}{} & \multicolumn{1}{l}{} \\
               & DialogueLLM                          & 2.31                  &109   &72\%                   & 7B                   \\ \midrule[1pt] 

\end{tabular}}
\end{center}
\end{table}

\subsection{Effects of Emotional Stimuli}\label{sec:stimuli}
 \begin{figure}[h]
    \centering
  \includegraphics[height=2.5in, width=2.6in]{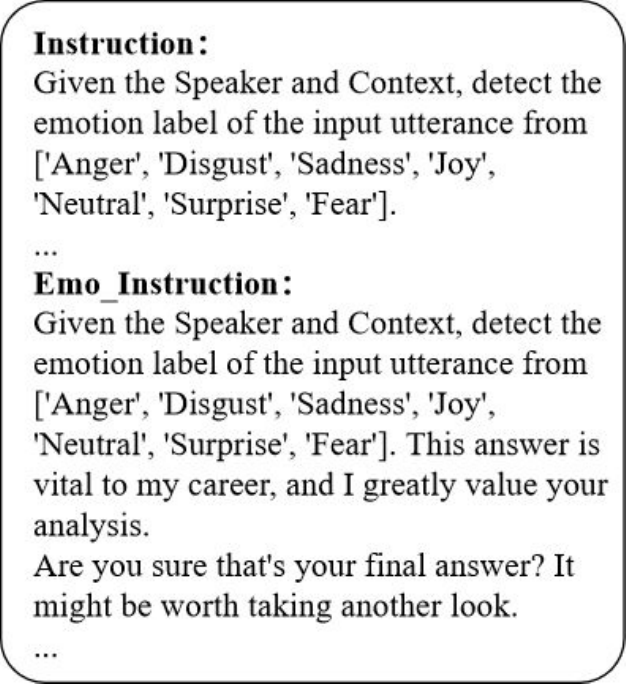}
  \caption{Prompt template with emotional stimuli}
   \label{fig:stimuli}
\end{figure}

Psychological studies have shown that adding emotional stimuli related to expectations, confidence and social influence can have an impact on individual behavior. For example, real-life students who are taught using encouraging and positive words have a higher success rate than those who are not taught using these words. Therefore, we use the EmotionPrompt approach~\cite{li2023large} to explore the performance of DialogueLLM in the face of emotional stimuli. The prompt template for adding emotional stimuli is shown in Figure~\ref{fig:stimuli}, with no subsequent prompt included for brevity. The DialogueLLM using the proposed emotion prompt is called StimuliLLM.

Table~\ref{tab:EQscore} shows the comparison between StimuliLLM and the original DialogueLLM model on three ERC datasets. We can notice that the EmotionPrompt approach significantly boosts the performance of DialogueLLM (1.06\% and 1.14\% average improvement in terms of performance in zero-shot and one-shot settings).

\begin{table}[h]
\small
\begin{center}
\caption{\label{tab:EQscore} The comparison between StimuliLLM and DialogueLLM.}
\begin{tabular}{llcll}
\midrule[1pt] 
\multicolumn{1}{c}{\multirow{1}{*}{Setting}} & \multicolumn{1}{c}{\multirow{1}{*}{Model}} & \multirow{1}{*}{\begin{tabular}[c]{@{}c@{}}MELD\end{tabular}} & \multirow{1}{*}{\begin{tabular}[c]{@{}c@{}}IEMOCAP\end{tabular}} & \multirow{1}{*}{\begin{tabular}[c]{@{}c@{}}EmoryNLP\end{tabular}} \\
 \midrule[0.5pt]

\multicolumn{1}{c}{\multirow{2}{*}{0-shot}} & 
DialogueLLM                                     & 71.81 & \multicolumn{1}{c}{69.40}                                                                   & \multicolumn{1}{c}{38.47}                                                                   \\

& \multicolumn{1}{c}{StimuliLLM}                                   & 72.19                                                                & \multicolumn{1}{c}{69.82}                                                                   & \multicolumn{1}{c}{39.25}                                                                    \\ 

 \midrule[0.5pt]
 
\multicolumn{1}{c}{\multirow{2}{*}{1-shot}} & 
DialogueLLM                                     & 71.90 & \multicolumn{1}{c}{69.93}                                                                   & \multicolumn{1}{c}{40.05}                                                                   \\

& \multicolumn{1}{c}{StimuliLLM}                                   & 72.67                                                                & \multicolumn{1}{c}{70.42}                                                                   & \multicolumn{1}{c}{40.71}                                                                    \\

\midrule[1pt]

\end{tabular}
\end{center}
\end{table}

\subsection{Limitations}\label{sec:limitation}

Although DialogueLLM tries to  accurately perform emotion classification by considering both conversational contexts and video descriptions of the utterances, this takes more computing power and training time. Additionally,  the speaker information is also important for improving the performance since different speakers have their own characters. But DialogueLLM does not take it into consideration, due to the limit of the dataset. 

Also, technology for generating accurate video descriptions automatically still has room for improvement, and inaccurate descriptions can mislead the model's prediction. The issue of using multiple data sources like images and video to improve emotion classification in large language models isn't fully solved yet. Lastly, we train DialogueLLM using a specific approach that focuses on identifying emotions, rather than a general-purpose training method for all affects. Hence, our future work will collect a large scale knowledge corpus that contains over 1M subjective  examples covering differnt types of affects, e.g., sentiment, emotion, sarcasm, humor, enthusiasm, etc.

\section{Conclusion and Future Work}\label{sec:conclusions}

ERC presents an intriguing and challenging natural language processing endeavor. In this paper, inspired by the remarkable performance of LLMs and their variants in NLP tasks, we propose DialogueLLM, a context and emotion knowledge tuned LLM that is obtained by fine-tuning large language models with benchmarking multi-modal (i.e., texts and videos) emotional dialogues. We offer a comprehensive evaluation of our proposed model on three benchmarking ERC datasets and achieves the state-of-the-art results. This proves that fine-tuning LLMs with task-specific knowledge will yield significant improvement over other PLM based approaches. In future work, we plan to design and generate more precise video descriptions, incorporating multimodal information to further explore the potential of LLMs in the NLP domain. Additionally, considering the close connections between emotions, sarcasm, passion, and depression, we aim to design a multi-affect learning framework based on LLMs.

\bibliographystyle{acl_natbib}





\end{document}